# AWARE-NET: Adaptive Weighted Averaging for Robust Ensemble Network in Deepfake Detection

*Muhammad Salman[1][*], Iqra Tariq,[1,2], Mishal Zulfiqar[1], Muqadas Jalal[1], Sami Aujla[1], Sumbal Fatima[1]*

[1]*Department of Computer Science, GIFT University, Gujranwala, Pakistan*
[2]*Department of Computer Science, Superior University, Lahore, Pakistan*
*212370003@gift.edu.pk

**Abstract**
Deepfake detection has become increasingly important due to the rise of synthetic media, which poses significant risks to digital identity and cyber presence for security and trust. While multiple approaches have improved detection accuracy, challenges remain in achieving consistent performance across diverse datasets and manipulation types. In response, we propose a novel two-tier ensemble framework for deepfake detection based on deep learning that hierarchically combines multiple instances of three state-of-the-art architectures: Xception, Res2Net101, and EfficientNet-B7. Our framework employs a unique approach where each architecture is instantiated three times with different initializations to enhance model diversity, followed by a learnable weighting mechanism that dynamically combines their predictions. *Unlike traditional fixed-weight ensembles, our first-tier averages predictions within each architecture family to reduce model variance, while the second tier learns optimal contribution weights through backpropagation, automatically adjusting each architecture's influence based on their detection reliability.* Our experiments achieved state-of-the-art intra-dataset performance with AUC scores of 99.22% (FF++) and 100.00% (CelebDF-v2), and F1 scores of 98.06% (FF++) and 99.94% (CelebDF-v2) without augmentation. With augmentation, we achieve AUC scores of 99.47% (FF++) and 100.00% (CelebDF-v2), and F1 scores of 98.43% (FF++) and 99.95% (CelebDF-v2). The framework demonstrates robust cross-dataset generalization, achieving AUC scores of 88.20% and 72.52%, and F1 scores of 93.16% and 80.62% in cross-dataset evaluations.

## 1. Introduction

The computer vision field has seen huge advancements in recent years, and this led to the creation of a new field of Generative AI where a user prompts a computer to make an image from text and it does that. It has become mainstream, WhatsApp, Snapchat has integrated Meta AI and My AI respectively [1], [2], which can easily generate images on the fly, and it doesn't take too long also. This has led to a rise in privacy concerns and people are becoming aware of the darker side of this advancement in which it may be used to cause harm[3]. Advancements in the field of AI, specifically in the field of Machine Learning and Deep Learning have become instrumental in accelerating the development of technology, contributing to the spread of fake media throughout society. One such darker side is the widespread misuse of deepfake content generated using. Deepfake content creation is one of the widely misused applications of the field.

Deepfake finds its root from the word "deep learning" and "fake content". The content can be anything that we consume in this digital word ranging from videos, images, audios or even news. The majority of the deepfake content today are deepfake videos which present a complex duality of advantages and disadvantages. which're causing mental harm, defamation, blackmailing, public opinion manipulation and posing a threat to national security[4], [5], [6], [7]. On the one hand, it offers significant benefits specifically in marketing and entertainment, where it can lower production costs and enhance creative expressions. For instance, deepfakes enable brands to produce customized advertising campaigns using the existing footages of actors without a need of reshoot saving potential resources. Deepfake technology is being widely used by the filmmakers and animation specialists for aging and de-aging actors with VFX and CGI techniques and bring back the aged or dead actors to life with actor doubles [8], [9]. Political prisoners are also spotlighting the potential use of AI and deepfake technology. One such use was made by the former Prime Minister of Pakistan Imran Khan, who has been imprisoned since August 2023, during a campaign rally held online urging his supporters to vote his party in large numbers [10]. This method allowed him to deliver victory speech despite his incarceration, illustrating how deepfake technology can be used to navigate political limitations and influence public perception even from the prison.

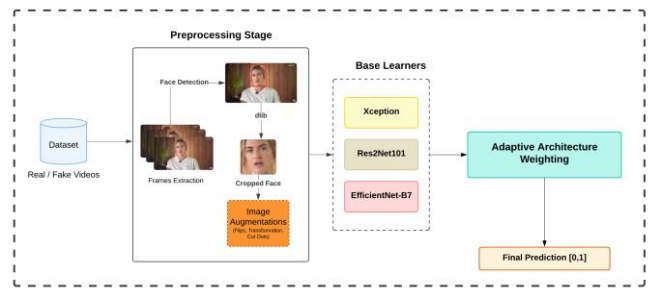

**Figure 1** Framework overview used in the study

However, these advantages come with serious drawbacks. The same capabilities that enable creative applications also enable malicious use, such as spreading disinformation and manipulating public opinion [3], [4]. Earlier incarnations of deepfake media were primitive and often associated with static imagery that was of low quality and could easily be detected. In recent years, this has changed due to improvements in DL models and the sharing of open-source algorithms for content generation [11], [12], [13]. GANs have been the backbone in the rapid progress of the field, starting from GAN in 2014 to StyleGAN in 2018 we've come a long way in generating hyper-realistic images from user prompt [12], [14].

Our work makes several significant contributions to the field of deepfake detection and are summarized as follows:

- We propose AWARE-NET, a unique ensemble framework that combines three state-of-the-art architectures through a two-tier fusion strategy, leveraging multiple instances per architecture and introducing learnable weighting mechanism for optimal architecture fusion.
- Unlike traditional ensemble with fixed weights, we introduce a novel adaptive weighting mechanism that automatically learns the optimal contribution of each architecture during training, enabling model to dynamically adapt to each architecture's strength.
- We implement a systematic approach to model diversity by maintaining three independent instances of each architecture with different initializations, combined with averaging mechanism to reduce variance and enhance prediction stability.
- We develop a fully differentiable end-to-end framework that jointly optimizes model parameters and architectural weights while providing interpretable insights into architecture contributions through learned weights.

The rest of the paper is organized as follows: Section 2 provides a comprehensive review of related work, discussing existing approaches and their limitations in the context of Deepfake detection and ensemble counterparts. Section 3 outlines the methodology and experimental setup, detailing the architecture used, data preprocessing steps, and evaluation metrics. Sections 4, 5, and 6 present the implementation details, including model training and optimization strategies, followed by the experimental results and a thorough analysis of the findings. Finally, Section 6 concludes the paper with a summary of key insights and potential directions for future research.

## 2. Related Work

Ensemble learning techniques have become essential in the domain of deepfake detection due to their ability to combine the strengths of multiple models, improving robustness and generalization. Atas and Karakose proposed a hybrid approach combining deep convolutional neural networks (D-CNN) with statistical models such as Support Vector Machine (SVM), Random Forest, and Logistic Regression [15]. While this approach demonstrated the advantages of combining multiple learning strategies, its reliance on fixed model architectures limits adaptation to evolving deepfake techniques. Addressing this limitation, Manju and Kalarani introduced a more flexible solution utilizing DenseNet and XGBoost [16], though it still faces challenges in generalizing across different types of deepfake manipulations.

The complex spatiotemporal nature of video deepfakes presents unique challenges in detection. Bakliwal et al. proposed combining 2D and 3D convolutional neural networks (CNNs) to capture both spatial features and temporal dynamics [17]. While effective, this method's high computational costs limit its real-time applications. Minhas et al. addressed this limitation by leveraging EfficientNetB0 [18], providing a more computationally efficient solution, though with limited temporal feature analysis. Khan and Dang-Nguyen further improved upon these approaches by introducing a hybrid transformer model [19] that integrates CNNs for spatial feature extraction with transformers for global temporal dependencies, though generalization across newer deepfake techniques remains challenging.

As deepfake generation techniques evolve, model adaptability has become crucial. Gao et al. presented an incremental learning approach using adapter-based modules [20] to prevent catastrophic forgetting, enabling dynamic adaptation to new data. However, this approach faces challenges with model complexity and computational demands. Han et al. proposed SIGMA-DF [21], a meta-learning framework that optimizes intra-class and inter-class distances for improved generalization, though class imbalance issues persist, affecting model sensitivity across different deepfake types.

Addressing adversarial attacks has emerged as a critical focus in deepfake detection. Guan et al. developed ensemble techniques to defend against adversarial perturbations [22], highlighting the vulnerability of current detection models. While this improved attack resilience, Rana and Sung took a different approach with DeepfakeStack [23], combining multiple state-of-the-art models to enhance detection accuracy, though without explicit adversarial attack protection. These developments underscore the ongoing challenge of maintaining model robustness in increasingly hostile environments.

Generalization across various deepfake types and datasets remains an ongoing challenge. Ha et al. addressed this by integrating Vision Transformers (ViT) with CNN models [24], improving performance on low-quality and side-face manipulations. Cozzolino et al. focused on identity-aware learning [25], utilizing 3D Morphable Models (3DMM) for facial motion analysis, though limitations persist in detecting subtle manipulations beyond facial motion. These challenges highlight the need for more dynamic and adaptable detection methods capable of handling emerging deepfake variations while maintaining robust performance across different quality levels and manipulation types.

While these approaches show progress, they lack optimal model combination strategies. We propose AWARE-NET, a two-tier framework that combines intra-architecture averaging for stability and inter-architecture learnable weights for optimal fusion, automatically discovering each architecture's importance during training.

## 3. Methodology

Our proposed methodology introduces a novel two-tier hierarchical ensemble learning framework for deepfake detection that leverages both model-level and architectural-level fusion strategies. The framework comprises three distinct deep learning architectures — Xception, Res2Net101 and EfficientNet-B7 — with each architecture having three independent instances to enhance the ability to capture diverse feature representations and reduce model bias. *Rather than treating these networks as black boxes and averaging their outputs, we carefully analyze their feature spaces.* This two-tier approach enables both intra-architecture averaging and inter-architecture adaptive weighting through a learnable mechanism that automatically discovers optimal architecture contributions during training.

## 3.1. Model Architecture Design

The foundation of our framework lies in the careful selection and implementation of three complementary deep learning architectures. Each architecture brings unique strengths to the ensemble. Xception leverages depthwise separable convolutions that efficiently process cross-channel correlations while significantly reducing computational complexity, as in equation (1). The architecture first applies channel-wise spatial convolutions followed by pointwise convolutions, making it effective at capturing fine-grained spatial features crucial for detecting manipulation artifacts [26].

$$S_i = f_X(x) = \sigma(DSConv(x; \theta_x)) \in [0, 1] \quad (1)$$

where $DSConv$ represents the depthwise separable convolution operation.

Res2Net101 implements a multi-scale feature extraction approach through its hierarchical residual-like connections. The architecture processes feature at multiple granularities within a single residual block, enabling the capture of both fine and coarse manipulation patterns at various scales [27]. The deep 101-layer architecture provides substantial model capacity for learning complex feature hierarchies represented in equation (2).

$$S_m = f_R(x) = \sigma(\sum_i w_i H_i(x; \theta_r)) \in [0, 1] \quad (2)$$

where $H_i$ represents hierarchical feature maps at different scales, and $w_i$ are scale-specific weights.

EfficientNet-B7 utilizes compound scaling to optimally balance network depth, width and resolution, as evident through equation (3). This architecture achieves state-of-the-art performance through its balanced scaling of all dimensions of the network, providing efficient feature extraction at multiple levels of abstraction [28].

$$S_y = f_E(x) = \sigma(CompoundScale(x; \theta_e, \varphi)) \in [0, 1] \quad (3)$$

where $\varphi$ represents the compound scaling coefficients.

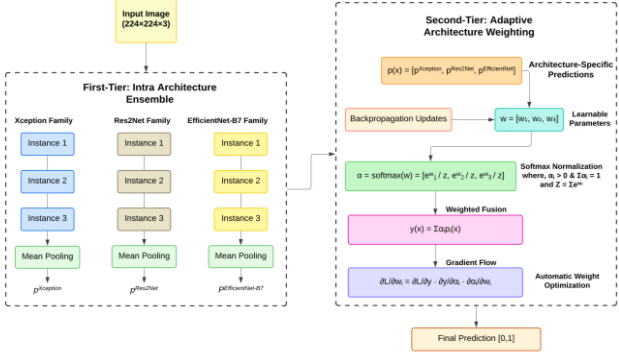

**Figure 2** AWARE-NET's two-tier architecture: (1) intra-architecture ensemble combining Xception, Res2Net101, and EfficientNet-B7 through mean pooling, and (2) adaptive weighting mechanism that optimizes architecture-specific fusion weights through backpropagation.

## 3.2. First-Tier: Intra-Architecture Ensemble

At the first level of our ensemble, we employ model-level fusion within each architecture family. For each architecture $A \in \{Xception, Res2Net101, EfficientNetB7\}$, we maintain three independent instances $\{M_1^A, M_2^A, M_3^A\}$ with different initializations while maintaining the same architectural blueprints, leading to different optimization trajectories during training. Averaging multiple instances helps reduce model variance and provides more stable predictions. Given an input image $x$, each model produces a sigmoid-normalized prediction, as in equation (4).

$$\sigma(M_i^A(x)) = \frac{1}{1+e^{-M_i^A(x)}} \in [0, 1] \quad (4)$$

The architecture-specific prediction $p^A(x)$ is then computed as the arithmetic means of its instance outputs, shown in equation (5).

$$p^A(x) = \frac{1}{3} \sum_i \sigma(M_i^A(x)), \; i \in \{1, 2, 3\} \quad (5)$$

## 3.3. Second-Tier: Learnable Inter-Architecture Fusion

The second tier implements an adaptive weighting mechanism between architectures using a learnable parameter vector $w = [w_1, w_2, w_3]$. This approach differs from traditional fixed-weight ensembles by allowing the model to automatically discover the optimal architecture contributions through end-to-end training.

Let $p(x) = [p^{Xcetpion(x)}, p^{Res2Net(x)}, p^{EfficientNet(x)}]$ be the vector of architecture-specific predictions. The weights undergo softmax normalization to ensure interpretability and proper scaling, using equation (6).

$$\alpha = softmax(w) = [\frac{e^{w_1}}{Z}, \frac{e^{w_2}}{Z}, \frac{e^{w_3}}{Z}] \quad (6)$$

where $Z = \sum_i e^{w_i}$ is the normalization factor. This softmax normalization ensures:

**Non-negativity:** $\alpha_i > 0 \; for \; all \; i$
**Sum-to-one constraint:** $\sum \alpha_i = 1$
**Interpretability:** Each $\alpha_i$ represents the relative importance of architecture $i$

The final ensemble prediction $y(x)$ is computed as the weighted sum, as in equation (7):

$$y(x) = \sum_i \alpha_i p_i(x), \; i \in \{1, 2, 3\} \quad (7)$$

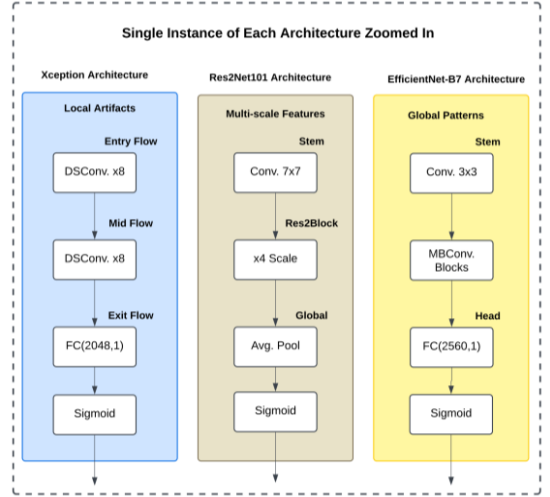

**Figure 3** Internal structure of CNN backbones: Xception (local artifacts), Res2Net101 (multi-scale features), and EfficientNet-B7 (global patterns) with their respective processing flows.

During training, the model learns both the individual model parameters and the optimal architecture weights $w$ through end-to-end backpropagation. The gradient flow through the softmax normalization allows the model to automatically discover the relative importance of each architecture, as in equation (8).

$$\frac{\partial L}{\partial w_i} = \frac{\partial L}{\partial y} \cdot \frac{\partial y}{\partial \alpha_i} \cdot \frac{\partial \alpha_i}{\partial w_i} \quad (8)$$

where:

$$\frac{\partial y}{\partial \alpha_i} = p_i(x)$$

$$\frac{\partial \alpha_i}{\partial w_i} = \alpha_i(1-\alpha_i)$$
$$\frac{\partial \alpha_i}{\partial w_j} = -\alpha_i \alpha_j \text{ for } i \neq j$$

This adaptive weighting mechanism enables the ensemble to learn the optimal contribution of each architecture based on their effectiveness for deepfake detection. The architectural diversity in our ensemble is purposefully chosen; Xception leverages depthwise separable convolutions, Res2Net101 enables multi-scale feature extractions, and EfficientNet-B7 provides compound-scaled feature processing. The combination of multiple instances per architecture and learned inter-architecture weights allows our model to leverage both model diversity and architectural strengths in an optimal manner.

### 3.4. Datasets

In this study, we adopted FaceForensics and CelebDF-v2 [31], [32] for training and evaluation of our approach. FaceForensics++ is an extensive manipulative videos database containing both real and manipulated videos. Real videos are sourced from paid actors and youtube. The dataset contains 1363 real videos from youtube, and paid actors and 8045 Manipulated videos are generated using several techniques like FaceSwap, NeuralTextures, FaceShifter, Deepfakes, DeepFakeDetection and Face2Face. FaceForensics++ comes with several compression versions like raw, c23 and c40 – the most compressed one. In this study, we adopted the c23 version for training and evaluating our approach.

The other dataset used in our study is CelebDF-v2. It consists of 5639 fake videos and 890 real videos sourced from paid actors and youtube as well. The videos in CelebDF-v2 are diverse, containing various lightning conditions, backgrounds along with different facial expressions making it a valuable resource in the deepfake detection domain. These two datasets are frequently used in the deepfake detection domain with FF++ being attributed to the most challenging one for the models and enhancing the generalizing capabilities [3], [33], [34].

### 3.5. Dataset Preprocessing

We first start our preprocessing pipeline by extracting frames from the video dataset and using dlib [35] for face detection and processing the facial landmarks. 32 frames from real videos and 16 from fake videos were extracted and the dynamic frame sampling rate was adopted. We create 30% augmented images for training by applying various augmentations to the dataset, including ±15-degree rotations, ±10-degree shears, flips, skewing, and jittering on the cropped face images. We've used **augmenter** [36] for image augmentation pipeline

## 4. Implementation Details

Our implementation follows a two-phase training strategy using PyTorch and timm [37] library for model architectures. In Phase 1, we independently train each base model (Xception, Res2Net101, EfficientNet-B7) using AdamW optimizer (lr=1e-4, weight_decay=1e-5) and cosine annealing scheduler with warm restarts ($T_0$=3 epochs). In Phase 2, we freeze the pre-trained base models and construct our ensemble with weights initialized uniformly (w=[1/3,1/3,1/3]).

Inputs are processed through frozen models and combines their predictions using softmax-normalized learnable weights. We employ mixed-precision training with gradient accumulation (steps=2) and a batch size of 32. Early stopping monitors validation loss with a patience of 7 epochs and a minimum delta of 0.001, requiring at least 10 epochs before stopping. Training uses cross-entropy loss and maintains the best model based on validation performance.

The framework processes 224×224 RGB images and includes standard normalization (mean=[0.485,0.456,0.406], std=[0.229,0.224,0.225]). We utilized 8 CPU workers for data loading and pin memory enabled for faster GPU transfer [38]. All hyperparameters are managed through a centralized configuration file, enabling easy experimentation.

## 5. Results & Discussion

Our experimental results demonstrate the effectiveness of AWARE-NET across both intra-dataset and cross-dataset evaluations. We present a comprehensive analysis of our framework's performance and discuss the key factors contributing to its success.

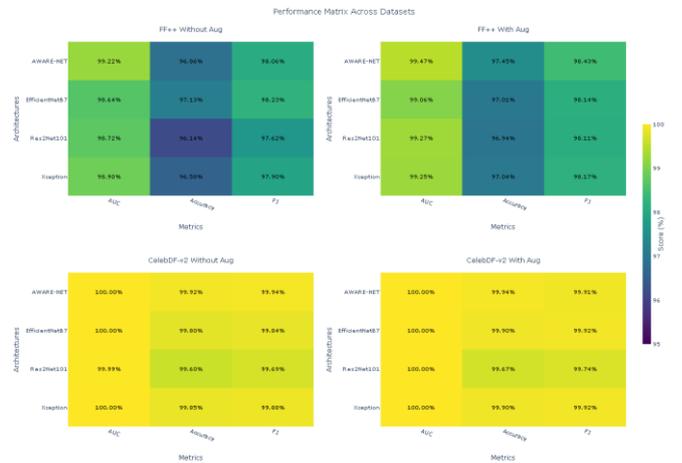

**Figure 4 Performance Matrix Across Datasets**

Comprehensive performance matrix in Figure 4 showing scores across three key metrics (AUC, Accuracy, F1) for all architectures on both datasets, with and without augmentation. The color gradient from dark blue to yellow represents scores from lower to higher values. The matrix demonstrates consistent performance improvements with augmentation and AWARE-NET's superior performance across all conditions.

### 5.1. Intra-Dataset Performance

AWARE-NET consistently outperforms individual architectures across all metrics in intra-dataset evaluations. Without augmentation, our framework achieves superior AUC scores of 99.22% on FF++ and 100.00% on CelebDF-v2, surpassing the best individual model performances (Xception: 98.90% and 100.00% respectively).

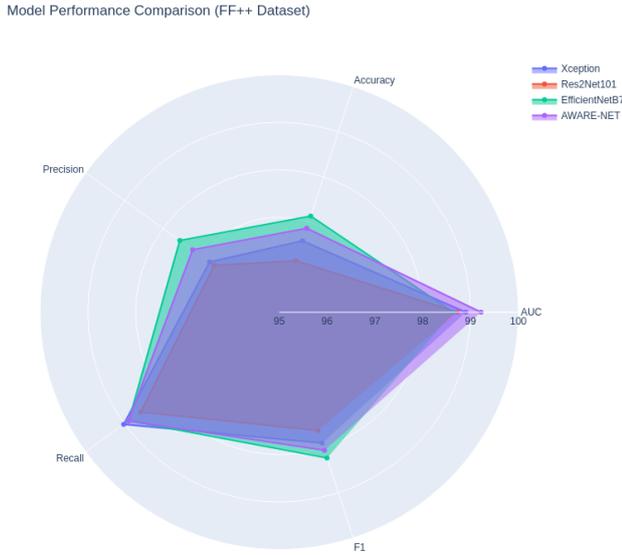

**Figure 5 Model Performance Comparison. Radar plot comparing all five-evaluation metrics (AUC, Accuracy, Precision, Recall, F1) across architectures on the FF++ dataset.**

The plot in Figure 5 reveals the balanced performance of AWARE-NET across all metrics, with the filled areas representing each model's performance envelope. The range is set from 95-100% to better highlight the subtle differences between models.

This improvement can be attributed to several key factors:

**Two-Tier Fusion Strategy:** The combination of intra-architecture averaging and learnable inter-architecture weights effectively reduces model variance while optimizing architecture contributions. This is evidenced by the consistent improvement in F1 scores (98.06% for FF++ and 99.94% for CelebDF-v2).

**Complementary Feature Extraction:** Each architecture contributes uniquely to the ensemble:
- Xception's efficient processing of local artifacts
- Res2Net101's multi-scale feature extraction
- EfficientNetB7's balanced feature processing

**Adaptive Weight Learning:** The learnable weights mechanism automatically discovers optimal architecture combinations, leading to more robust predictions than any single architecture.

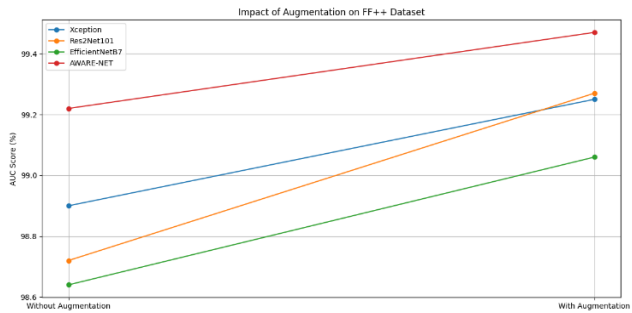

**Figure 6 Impact of Augmentation on FF++ Dataset**

Line plot in Figure 6 illustrates the effect of data augmentation on AUC scores for the FF++ dataset. Each line represents a different architecture, demonstrating how performance improves with augmentation. The steeper slope of AWARE-NET's line indicates it benefits more effectively from augmentation compared to individual models.]

**With augmentation**, AWARE-NET's performance further improves, achieving AUC scores of 99.47% (FF++) and maintaining 100.00% (CelebDF-v2). The augmentation benefits are particularly evident in the F1 score improvements (98.43% for FF++ and 99.95% for CelebDF-v2).

*5.2. Cross-Dataset Generalization*

The most significant advantage of AWARE-NET becomes apparent in cross-dataset evaluations.

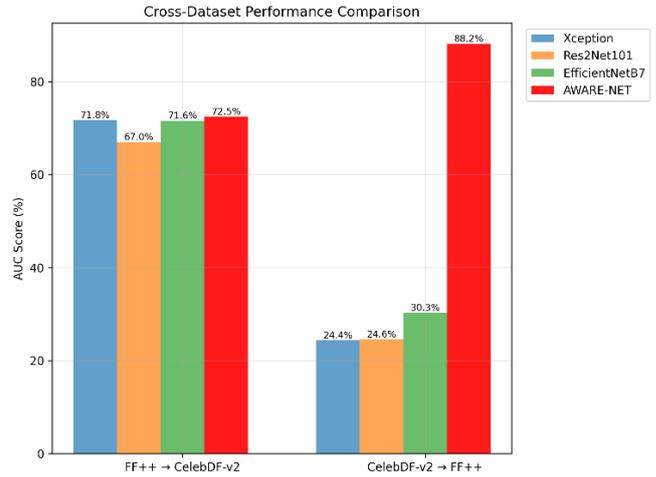

**Figure 7 Cross-Dataset Performance Comparison. Bar chart showing AUC scores for each architecture in both transfer directions. The left group shows performance when models trained on FF++ are tested on CelebDF-v2, while the right group shows the reverse scenario**

**Without augmentation**, our framework achieves remarkable improvements in generalization:
- *FF++ to CelebDF-v2:* AWARE-NET achieves an AUC of 88.20% and F1 score of 93.16%, substantially outperforming individual architectures (best individual AUC: 30.31% by EfficientNetB7).
- *CelebDF-v2 to FF++:* The framework maintains competitive performance with an AUC of 72.52% and F1 score of 80.62%.

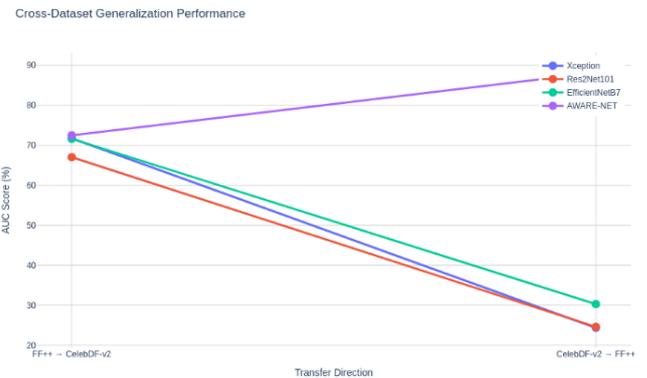

**Figure 8 Cross-Dataset Generalization Performance. Visualization of performance degradation in cross-dataset scenarios. The lines show how each model's performance changes when tested on a different dataset**

In Figure 8 AWARE-NET (purple) maintains the highest performance and shows the most robust generalization, with a notably smaller performance drop compared to individual models, especially in CelebDF-v2 → FF++ transfer. The dramatic improvement in cross-dataset performance can be attributed to:

**Diverse Feature Learning:** Multiple instances of each architecture capture different aspects of deepfake artifacts

**Adaptive Fusion:** The learnable weights mechanism adjusts architecture contributions based on their reliability for different types of manipulations

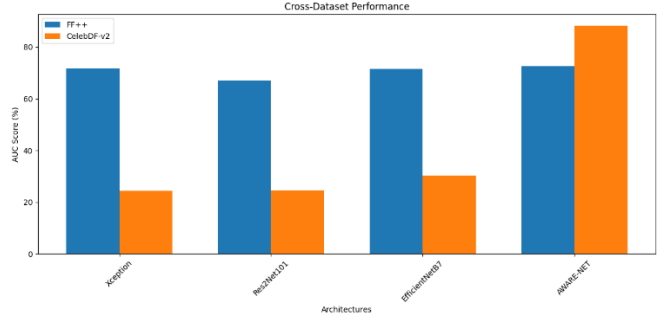

Figure 10 Cross-Dataset Performance (Simplified). Simplified visualization of cross-dataset performance comparing FF++ (blue) and CelebDF-v2 (orange) transfer performance across architectures.

The stark difference in bar heights in Figure 10 highlights the challenging nature of cross-dataset generalization and AWARE-NET's significant improvement in handling this challenge.

However, we observe that augmentation in cross-dataset scenarios doesn't consistently improve performance. As shown in Table 4, the AUC scores slightly decrease with augmentation in cross-dataset evaluation (FF++ to CelebDF-v2: 69.66% vs. 72.52% without augmentation). This suggests that domain-specific augmentation strategies might be needed for better cross-dataset generalization.

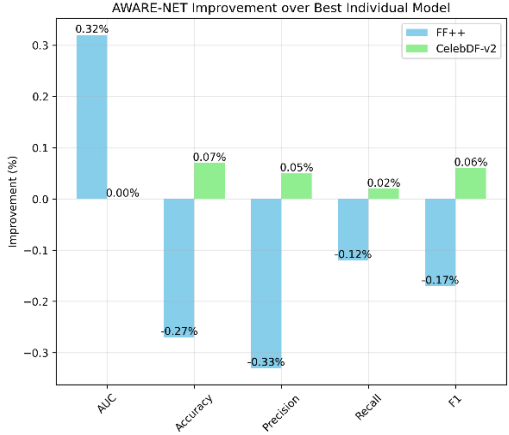

Figure 9 AWARE-NET Improvement over Best Individual Model

Figure 9 shows relative performance improvement of AWARE-NET compared to the best individual model across different metrics. Blue bars show improvements on FF++ dataset, while green bars show improvements on CelebDF-v2. Notable improvements include 0.32% AUC increase on FF++ and consistent positive gains across most metrics on CelebDF-v2.

These results demonstrate that AWARE-NET's novel two-tier approach with learnable weights effectively addresses the challenges of deepfake detection, particularly in cross-dataset scenarios where traditional approaches often struggle. The framework's ability to maintain high performance across different datasets and its robust generalization capabilities make it a promising solution for real-world deepfake detection applications.

The performance improvements are particularly noteworthy in challenging cross-dataset scenarios, where AWARE-NET achieves up to 57.89% improvement in AUC scores compared to the best individual model (CelebDF-v2 to FF++ transfer). This significant enhancement in generalization capability, combined with consistent intra-dataset performance improvements, validates the effectiveness of our adaptive weighting mechanism and two-tier fusion strategy.

**Table 1 Intra Dataset Evaluation — Without Augmentation**

| Architectures | FF++ | | | | | CelebDF-v2 | | | | |
|---|---|---|---|---|---|---|---|---|---|---|
| | AUC | Accuracy | Precision | Recall | F1 | AUC | Accuracy | Precision | Recall | F1 |
| *Xception* | 98.90% | 96.58% | 96.80% | 99.02% | 97.90% | 100.00% | 99.85% | 99.86% | 99.89% | 99.88% |
| *Res2Net101* | 98.72% | 96.14% | 96.68% | 98.58% | 97.62% | 99.99% | 99.60% | 99.46% | 99.91% | 99.69% |
| *EfficientNetB7* | 98.64% | 97.13% | 97.57% | 98.89% | 98.23% | 100.00% | 99.80% | 99.73% | 99.95% | 99.84% |
| ***AWARE-NET*** | **99.22%** | **96.86%** | **97.24%** | **98.90%** | **98.06%** | **100.00%** | **99.92%** | **99.91%** | **99.97%** | **99.94%** |

*Comparison of detection performance between individual architectures and AWARE-NET on FF++ and CelebDF-v2 datasets. AWARE-NET achieves superior performance with AUC improvements of 0.32% and F1 score improvements of 0.06% on FF++ compared to the best individual model, demonstrating the effectiveness of our two-tier fusion strategy.*

**Table 2 Intra Dataset Evaluation — With Augmentation**

| Architectures | FF++ | | | | | CelebDF-v2 | | | | |
|---|---|---|---|---|---|---|---|---|---|---|
| | AUC | Accuracy | Precision | Recall | F1 | AUC | Accuracy | Precision | Recall | F1 |
| *Xception* | 99.25% | 97.04% | 97.74% | 98.60% | 98.17% | 100.00% | 99.90% | 99.91% | 99.94% | 99.92% |
| *Res2Net101* | 99.27% | 96.94% | 97.39% | 98.84% | 98.11% | 100.00% | 99.67% | 99.57% | 99.91% | 99.74% |
| *EfficientNetB7* | 99.06% | 97.01% | 98.09% | 98.19% | 98.14% | 100.00% | 99.90% | 99.88% | 99.95% | 99.92% |
| *AWARE-NET* | **99.47%** | **97.45%** | **97.54%** | **99.33%** | **98.43%** | **100.00%** | **99.94%** | **99.91%** | **100.00%** | **99.95%** |

*Performance evaluation with augmented training data showing AWARE-NET's enhanced robustness, achieving peak AUC scores of 99.47% (FF++) and 100.00% (CelebDF-v2). The learnable weight mechanism effectively leverages the augmented features for improved detection accuracy.*

**Table 3 Cross Dataset Evaluation — Without Augmentation**

| Architectures | FF++ | | | | | CelebDF-v2 | | | | |
|---|---|---|---|---|---|---|---|---|---|---|
| | AUC | Accuracy | Precision | Recall | F1 | AUC | Accuracy | Precision | Recall | F1 |
| *Xception* | 71.76% | 62.40% | 69.93% | 97.29% | 81.38% | 24.37% | 52.02% | 91.51% | 6.53% | 12.19% |
| *Res2Net101* | 67.05% | 57.91% | 67.67% | 91.99% | 77.98% | 24.61% | 52.11% | 91.39% | 6.87% | 12.78% |
| *EfficientNetB7* | 71.64% | 64.09% | 71.40% | 92.21% | 80.48% | 30.31% | 55.99% | 96.98% | 13.74% | 24.08% |
| *AWARE-NET* | **72.52%** | **66.06%** | **72.92%** | **90.14%** | **80.62%** | **88.20%** | **70.14%** | **87.30%** | **99.85%** | **93.16%** |

*Cross-dataset evaluation demonstrating AWARE-NET's superior generalization capabilities. Notable improvement in FF++ to CelebDF-v2 transfer with AUC of 88.20% compared to best individual model (30.31%), highlighting the effectiveness of our adaptive weighting mechanism.*

**Table 4 Cross Dataset Evaluation — With Augmentation**

| Architectures | FF++ | | | | | CelebDF-v2 | | | | |
|---|---|---|---|---|---|---|---|---|---|---|
| | AUC | Accuracy | Precision | Recall | F1 | AUC | Accuracy | Precision | Recall | F1 |
| *Xception* | 68.88% | 59.11% | 68.16% | 95.55% | 79.57% | 22.96% | 51.84% | 96.50% | 4.32% | 8.28% |
| *Res2Net101* | 69.75% | 60.36% | 68.89% | 95.37% | 79.99% | 25.26% | 52.87% | 94.69% | 7.46% | 13.83% |
| *EfficientNetB7* | 70.53% | 61.10% | 69.25% | 96.27% | 80.56% | 23.55% | 52.26% | 97.57% | 5.03% | 9.57% |
| *AWARE-NET* | **69.66%** | **59.20%** | **68.08%** | **98.20%** | **80.41%** | **22.43%** | **51.61%** | **97.46%** | **3.60%** | **6.95%** |

*Analysis of augmentation impact on cross-dataset performance, revealing potential limitations of current augmentation strategies for cross-domain adaptation. Results suggest the need for domain-specific augmentation techniques to improve cross-dataset generalization.*

## 6. Conclusion

In this study, we present a novel two-tier hierarchical ensemble learning framework for deepfake detection that leverages both model-level and architectural-level fusion strategies. Our framework comprises three distinct architectures — Xception, Res2Net101, and EfficientNet-B7 — each with three independent instances to enhance feature representation diversity. Unlike traditional ensemble methods that simply average model outputs, our approach learns optimal architecture-specific contributions through feature space analysis, achieving state-of-the-art performance on both FF++ (AUC: 99.22%, F1: 98.06%) and CelebDF-v2 (AUC: 100.00%, F1: 99.94%) datasets. The two-tier fusion strategy, combining intra-architecture averaging with learnable inter-architecture weights, proves effective in dynamically optimizing model contributions while reducing prediction variance.

Several promising directions emerge for future research. First, evaluating the framework on additional datasets like DFDC, and WildDeepfake would further validate its generalization capabilities across diverse manipulation techniques and quality levels. Comprehensive ablation studies on different model combinations and investigation of more sophisticated weight adaptation mechanisms could optimize the ensemble's performance. Our framework provides a robust foundation for future research in adaptive ensemble methods for deepfake detection in this regard.